\documentclass[11pt,letterpaper]{article}
\usepackage{cogsys}
\usepackage[T1]{fontenc}
\usepackage{times}
\usepackage[pdftex]{graphicx} % use this when importing PDF files
\usepackage{amsmath}

\usepackage{natbib}
\usepackage{url}

\cogsysheading{X}{2022}{1-19}{10/2022}{11/2022}
\ShortHeadings{Inducing Language Models Via Concept Formation}
              {C.J.\ MacLellan, P.\ Matsakis, and P.\ Langley}

\begin{document} 

\title{Efficient Induction of Language Models Via\\ Probabilistic Concept Formation}
 
\author{Christopher J. MacLellan}{cmaclell@gatech.edu}
\address{Teachable AI Lab, Georgia Institute of Technology, 
         Atlanta, GA 30332 USA}
\author{Peter Matsakis}{petermatsakis@gmail.com}
\author{Pat Langley}{patrick.w.langley@gmail.com}
\address{Institute for the Study of Learning and Expertise, 
         2164 Staunton Court, Palo Alto, CA 94306 USA}
\vskip 0.2in
 
\begin{abstract}
This paper presents a novel approach to the acquisition of language models from corpora. The framework builds on Cobweb, an early system for constructing taxonomic hierarchies of probabilistic concepts that used a tabular, attribute-value encoding of training cases and concepts, making it unsuitable for sequential input like language.  In response, we explore three new extensions to Cobweb---the Word, Leaf, and Path variants.  These systems encode each training case as an anchor word and surrounding context words, and they store probabilistic descriptions of concepts as distributions over anchor and context information. As in the original Cobweb, a performance element sorts a new instance downward through the hierarchy and uses the final node to predict missing features. Learning is interleaved with performance, updating concept probabilities and hierarchy structure as classification occurs.  Thus, the new approaches process training cases in an incremental, online manner that it very different from most methods for statistical language learning.  We examine how well the three variants place synonyms together and keep homonyms apart, their ability to recall synonyms as a function of training set size, and their training efficiency.  Finally, we discuss related work on incremental learning and directions for further research. 
\end{abstract}

%We report the system's ability to predict masked words in novel test
% sentences, showing that it achieves performance comparable to 
% mainstream systems. We also show that it requires substantially 
% less training data and processing time than these techniques.
% In closing, we outline ways to extend the framework to handle 
% other language-related tasks.

\section{Introduction}

The past decade has seen substantial progress in the development 
of practical language modeling approaches.
For example, the influential Word2Vec system \citep{mikolov2013efficient,mikolov2013distributed} demonstrated that one 
can extract meaningful semantic information, in the form of low-dimensional 
word embeddings, by analyzing words and their surrounding contexts.
The increased availability of massive training corpora, inexpensive memory 
storage, and very rapid computing abilities has fueled transformation 
of this early approach into sophisticated techniques for learning 
large language models, which are now used widely to extract and 
store semantic content from text. 

Although these systems are widely viewed as successful, they have 
some important drawbacks. 
First, they require {\it substantial} training data and computational 
resources. 
For example, the GPT-3 system was trained on a giant corpus that represents a sizable portion of the internet \citep{Brown2020gpt}. 
Some estimates suggest that its computing costs for training 
were as high as \$12 million \citep{wiggers_2020}.
In addition, systems like GPT utilize batch learning, which 
means that they train on all available data at once and thus cannot 
update their model efficiently in light of new cases. 
Furthermore, \citet{french1999catastrophic} has shown that neural 
network approaches, which include Word2Vec and GPT, often  
experience catastrophic forgetting---where they lose access to 
the results of early learning if given new cases. 
As a result, when new instances become available, the systems must 
be retrained on both old and new cases, which adds to development
and maintenance costs.

Recent efforts have tried to offset these expenses by exploring how language models that are trained on large corpora might be subsequently ``fine tuned'' using data from a target domain \citep{devlin2018bert}.
This approach aims to boost accuracy on the target task by extracting and transferring general knowledge from a large corpus to the target, while lowering overall costs by reusing a previously trained model.
Despite the excitement around large pre-trained language models, a recent study by Krishna et~al.\ (\citeyear{krishna2022}) suggests that little knowledge actually transfers through fine tuning; their work suggests that the main factor that aids performance is data from the target domain.\footnote{Specifically, their experiments show that language models created using data from the target domain for both pre-training and tuning (with no pre-trained model) achieve comparable performance to ones constructed by taking a pre-trained language model and then tuning it in the target setting.}
By only using target domain data, this approach reduces the data needed to build a language model by factors from 10 to 500, but it still requires large training sets to achieve reasonable performance.
Regardless of the pre-training scheme used, if new data becomes available regularly and models are repeatedly fine tuned with new data, then they are still likely to encounter catastrophic forgetting.

These challenges present major barriers to widespread adoption of such learning technology.
To overcome these drawbacks, we propose a new approach to language 
induction that takes inspiration from human learning.
We aspire to build a new class of induction systems that adhere to 
the constraints \citet{langley2022computational} has enumerated, in 
particular that they acquire {\bf modular} structures in a {\bf piecemeal} 
and {\bf incremental} way that builds on {\bf prior knowledge} to guide 
learning, so they can acquire expertise {\bf rapidly} from reasonably 
few training cases.
There has been some research into systems that adhere to these constraints 
(e.g., Mitchell et~al., \citeyear{mitchell2018never}), but the area 
deserves far more attention, especially considering the challenges 
faced by popular language learning techniques.

The basis for our response is Fisher's (\citeyear{fisher1990knowledge})
Cobweb, an early system for unsupervised learning of probabilistic 
concept hierarchies that was inspired by psychological findings on 
human categorization. 
In the next section, we review the framework's core assumptions about 
representation, performance, and learning. 
Next we present three adaptations of Cobweb that let it process words 
and their surrounding contexts, which we will see differ mainly in 
their representation of learned categories. 
After this, we report empirical studies of the variants' behaviors, 
focusing on their ability to recognize synonyms and distinguish 
homonyms but also examining their computational complexity. 
We conclude by discussing related work and outlining priorities 
for future research. 

% This system categorizes new instances down a taxonomic hierarchy and 
% uses terminal nodes in the hierarchy to generate predictions about 
% the instance.
% When the system is learning from a new instance, it engages in a 
% similar categorization process where it creates, deletes, and 
% updates intermediate nodes in the hierarchy.
% By leveraging its taxonomic hierarchy, Cobweb is able to efficiently
% update its knowledge when it encounters new experiences, without 
% having to reprocess all of its previous examples.

% Although Cobweb supports all of the human-like learning constraints 
% we outlined, it does not support the ability to process sequential, 
% textual inputs.
% It is also unclear how best to represent and reason about word context 
% within the Cobweb framework.
% This paper aims to address these issues through the exploration of 
% three new variants of the Cobweb model. 

%The first model directly translates Word2Vec ideas into the Cobweb framework.
%The other two models extend the first model to explore new ways of representing and reasoning about word context.

%Across this work, we aim to support three broad claims: (1) it is possible to support more efficient incremental language model acquisition by fusing Cobweb and Word2Vec ideas, (2) our initial models show promise for correctly grouping synonyms and homonyms. 

\section{A Brief Review of Cobweb}

\begin{figure}[t!]
    \centering
    \includegraphics[width=1.0\textwidth]{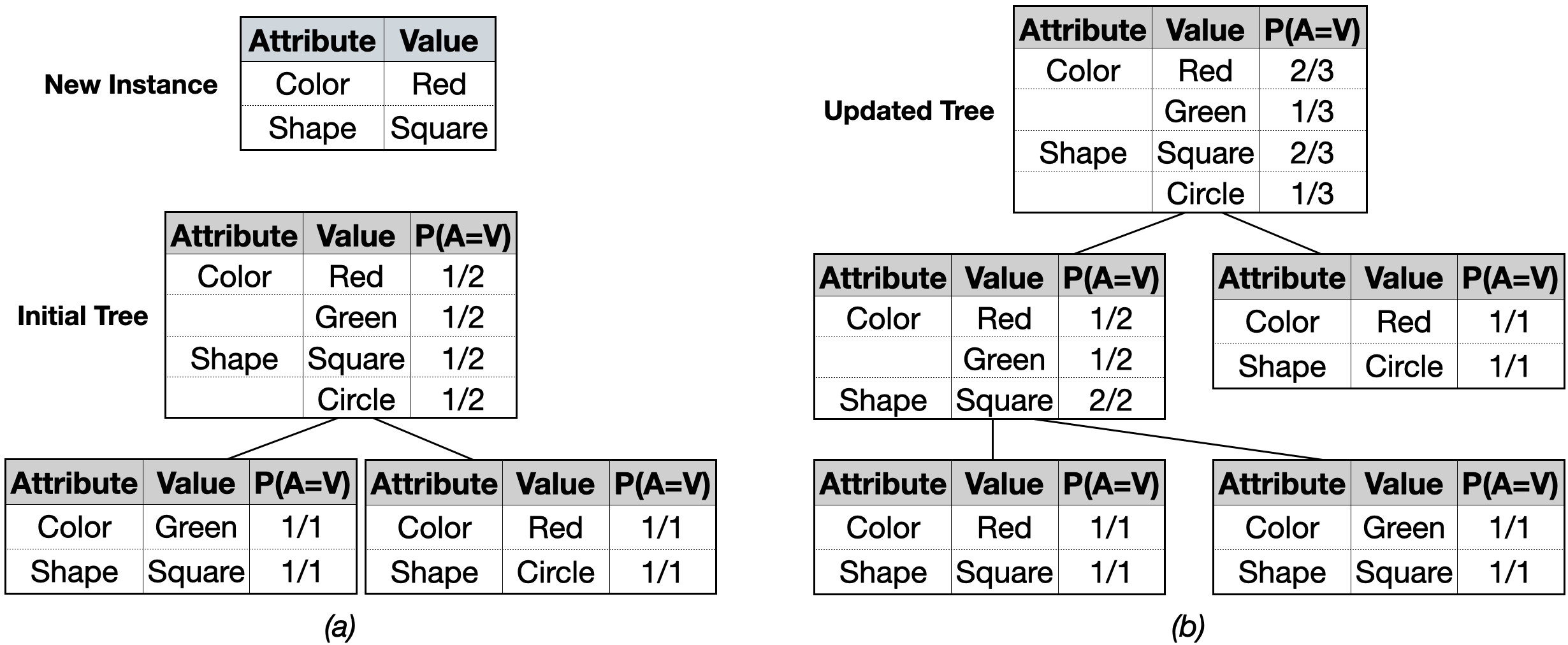}
    \vskip -0.07in
    \caption{An example of how an initial Cobweb hierarchy (a) is updated (b) after incorporating the instance shown at the top of (a). Each instance is described by attributes and their associated values. Concepts extend this representation to maintain attribute-value counts, which can be used to compute the probability of each attribute value conditioned on the category.}
    \label{fig:cobweb-example}
\end{figure}

Cobweb \citep{fisher1990knowledge} is an early system for unsupervised 
learning that constructs a taxonomic hierarchy of {\it concepts} 
from {\it instances} that are presented sequentially, incorporating 
insights from analyses of human categorization \citep{corter1992explaining}.
Both instances and concepts are described by tables of attribute values.
Instances assign a single value to each attribute, as seen in the top of 
Figure \ref{fig:cobweb-example} (a), which shows an instance for a red square.
In contrast, concepts store multiple possible values for each attribute,  
along with counts that specify the probabilities for different  
attribute values, as seen in the figure's other tables.
This {\it probabilistic} concept representation makes Cobweb robust 
in the face of uncertainty and noise. 
Equally important is the ability to store categories at different 
levels of generality and to organize them in a taxonomic hierarchy, 
which has major implications for performance and learning. 

% distinguishes Cobweb from other concept formation systems, such as 
% SAGE \citep{mclure2010learning}, that maintain a set of deterministic 
% features to describe each concept. Leveraging this representation, 
% makes the system

% Cobweb's core performance task is to predict missing attribute values 
% in new instances.
% Unlike supervised systems that are only capable of predicting a 
% single privileged attribute, this system is unsupervised, so it 
% is capable of predicting any of the attributes.
% To make predictions, an instance is sorted down the categorization tree. 

The basic performance mechanism involves sorting a new instance 
downward through the probabilistic concept hierarchy. 
Starting at the root note, Cobweb recursively considers whether to 
direct the case to one of the children or whether to stop at the 
current node. 
When this process terminates, it uses the final concept's probability 
table to predict the values for any missing attributes.
For example, imagine we have an instance whose shape is a circle 
and we want to predict its color. 
To sort this case through the concept hierarchy shown in Figure \ref{fig:cobweb-example}(b), we would start at the root and find 
that the rightmost child best describes it.
This child does not have any children, so categorization would halt 
and Cobweb would predict red as the color based on the stored 
probability table. 

During learning, Cobweb follows a similar process, but it updates the 
counts stored at a given concept node to reflect attribute values 
of the new instance assigned to it.
In addition to considering {\bf adding} the instance to each existing child concept, the system also considers three restructuring operations: 
% The system also considers four restructuring operations: {\bf adding} the instance to an existing child concept; 
{\bf creating} a new 
concept as a child of the current one; creating a new child that 
{\bf merges} two concepts that best describe the instance and then 
adding it to this new merged node, with the two existing concepts
becoming its children; and {\bf splitting} a child and promoting 
its children to become children of the current node and repeating 
the categorization process at the current node. 
These operations let Cobweb update its conceptual organization to 
better reflect new instances that it encounters and mitigate the  
effects of training order, which can influence the hierarchy structure. 

As an example, suppose a new instance (a red square) is added to 
the hierarchy in Figure \ref{fig:cobweb-example}(a). 
Cobweb starts at the root and updates its counts to include the 
instance description.
Next, it determines that the leftmost child (representing green 
squares) best matches the case.
Because this child is a leaf, it incorporates the instance into 
it by creating a new node that combines counts from the 
leaf with ones from the new case. 
The system then stores two children of this category---the previous 
leaf and a new concept based on the instance.
This process yields the updated concept hierarchy in Figure \ref{fig:cobweb-example}(b).
One interesting observation is that leaves of the tree always denote 
unique instances, whereas intermediate nodes encode probabilistic 
summaries of instances that fall beneath them.
By retaining these instances in the hierarchy, Cobweb can efficiently 
reorganize its knowledge in light of new experience without having 
to review all previous training cases.

During both performance and learning, Cobweb must decide which 
operations to carry out during sorting. 
To guide decision making, it uses {\it category utility} 
\citep{fisher1990knowledge,corter1992explaining}, a measure 
related closely to mutual information and similar to the 
information-gain metric used in decision-tree induction 
\citep{quinlan1986induction}. The function is defined as:

\begin{equation*}
    \frac{\sum^n_{k=1}P(C_k) \left[ \sum_i \sum_j P(A_i=V_{ij}|C_k)^2 - \sum_i \sum_j P(A_i=V_{ij})^2 \right]}{n} ~ .
\end{equation*}

%\begin{equation*}
%    EC(C) = \sum_{A \in C_{anchors}} \left[P(A|C)^2\right] + \sum_{L \in C_{conctext}} \left[P(L|C)^2\right] 
%\end{equation*}

\noindent
In plain terms, the $\sum_i \sum_j P(A_i=V_{ij}|C_k)^2$ term denotes 
the number of attributes that we expect to guess correctly for 
a given child $C_k$, assuming the system guesses an attribute value 
according to observed probabilities stored in $C_k$ and this guess 
is correct according to that same probability.
Similarly, the $\sum_i \sum_j P(A_i=V_{ij})^2$ term refers to the 
expected number of correct guesses in the parent.
The numerator represents the average increase in the number of attribute 
values correctly guessed in the children versus the parent, weighted 
by the probability of each child. 
The function divides this score by the number of children ($n$), 
which lets it compare candidates with different numbers of children 
(e.g., adding to a node vs.\ creating a new node).
To use category utility for decision making, Cobweb simulates each 
operation, computes the metric over resulting decompositions, and
selects the alternative that yields the highest score, with ties broken 
randomly.

The learning process is inherently efficient because it makes uses of
a tree structure.
\citet{fisher1990knowledge} shows that it takes $O(B^2 \times log_B(n) \times AV)$ steps to incorporate a new instance into the hierarchy, where $B$ is the average branching factor of the tree, $n$ is the number of previously classified instances, and $AV$ is the number of attribute values that appear across these instances.\footnote{Our Cobweb implementation includes an optimization that reduces this to $O(B \times log_B(n) \times AV)$, but we  have not yet published this result, so we include the higher estimate here.}
When applied to processing a set of $N$ instances incrementally, the 
overall run time becomes $O(N \times B^2 \times log_B(N) \times AV)$.
These computational characteristics offer support for efficient on-line  
learning, making the approach worth renewed attention, especially for 
large training sets. 

In summary, Cobweb provides a rich framework for investigating different 
aspects of concept formation, which has led to many extensions and variants.
For example, Cobweb/3 \citep{mckusick1990cobweb} extends the original framework, which only supports nominal attributes, to handle numeric attributes by assuming normal distributions over their values.
Other variants, such as Labyrinth \citep{thompson1991concept} and 
Trestle \citep{maclellan2016trestle}, have explored extensions 
that support structured, relational representations through the use 
of structure-mapping mechanisms.
A more recent system, Convolutional Cobweb \citep{maclellan2022convolutional}, combines the framework with convolutional processing to learn visual concepts. 
Iba and Langley (\citeyear{iba2011cobweb}) review a variety of other efforts 
that fall within this paradigm. 

% engage in the kinds of continual learning we discussed earlier, where incorporating a new instance does not require revisiting all prior training instances.

% From a psychological perspective, Cobweb unifies the exemplar and prototype theories \cite{smith2013categories}. 

\section{Contextual Extensions to Cobweb}

Despite the diversity of the Cobweb family and its many attractive features, no previous work has explored using it to acquire statistical language models. 
Our approach to adapting the system to this task takes inspiration from 
how Word2Vec \citep{mikolov2013efficient} captures contextual word 
regularities. 
Briefly, this system learns mappings from single words to points in a low-dimensional 
embedding space that, ideally, project words with similar meanings to 
nearby points. 
The semantics of a target or {\it anchor} word are defined with respect 
to its {\it context}, that is, the words that appear in its vicinity. 
In operational terms, an anchor's context is a window of surrounding words, 
such as the four words before it and the four words after it.

There are two schemes for inducing word embeddings with Word2Vec: Contextual Bag-of-Words (CBOW) and Skip-Grams.
The CBOW approach aims to find an embedding over the context words 
that predicts the anchor word.
In contrast, the Skip-Gram framework learns an embedding over  
anchor words that is useful for predicting words that appear 
in their context.
As per the bag-of-words label, the order of words in the context
window does not matter to CBOW.
This does not matter much for Skip-Gram, either, although it slightly 
favors choices that better predict words close to anchors. 
\citet{mikolov2013efficient} shows that both approaches produce 
viable embeddings, with CBOW faring better on some criteria (e.g., 
syntactic accuracy) and Skip-Gram better on others (e.g., semantic 
accuracy).
In this section, we explore how to incorporate ideas from both CBOW 
and Skip-Gram into Cobweb.
One important limitation of Word2Vec is that it maps every word 
to a single point in the embedding space. This means that it 
cannot represent different {\it senses} of a word, which is 
ironic given that these are closely linked to context. 

\citet{mikolov2013efficient} show that both approaches require $O(N \times E \times Q)$ steps during training, where $N$ is the number of words in the training corpus, $E$ is the number of training epochs used,  and $Q$, which is specific to each approach, captures the time to process a single anchor and its context. 
For CBOW, $Q$ is $W \times D + D \times log_2(V)$, where $W$ is the number of words in the context window, $D$ is the number of dimensions in the embedding, and $V$ is the size of the vocabulary (the number of unique words).
For Skip-Gram, $Q$ is $W \times (D + D \times log_2(V)$, where $W$ is the maximum distance of context words from the anchor that are used for prediction, a window that is typically slightly larger than the W term used in the CBOW approach.
If we use Word2Vec for on-line learning, then it will be far less 
efficient because we must retrain it on all the words after encountering 
each new training case. 
Used in this way, the training time for Word2Vec becomes $O(N^2 \times E \times Q)$.

Analysis of this earlier research suggested three different approaches 
to learning distributional knowledge about language  with Cobweb, which 
we will refer to as the {\bf Word}, {\bf Leaf}, and {\bf Path} variants. 
As we discuss in the remainder of the section, the first combines 
ideas from both Word2Vec techniques within the framework of probabilistic 
concept hierarchies. 
The other variants build on this initial scheme and introduce progressively 
more sophisticated ways to incorporate information about word contexts 
during both performance and learning. 

\subsection{The Cobweb Word System} 

As noted above, the CBOW version of Word2Vec learns embeddings for predicting the anchor word given its context, whereas the Skip-Gram approach finds embeddings for predicting the context words given the anchor. 
Our first Cobweb variant learns a concept hierarchy that optimizes jointly for prediction of both anchor and context words---effectively combining the CBOW and Skip-Gram criteria. 
This requires changes to Cobweb's structures, particularly its representation for instances. 

\begin{figure}[t!]
    \centering
    \includegraphics[width=0.8\textwidth]{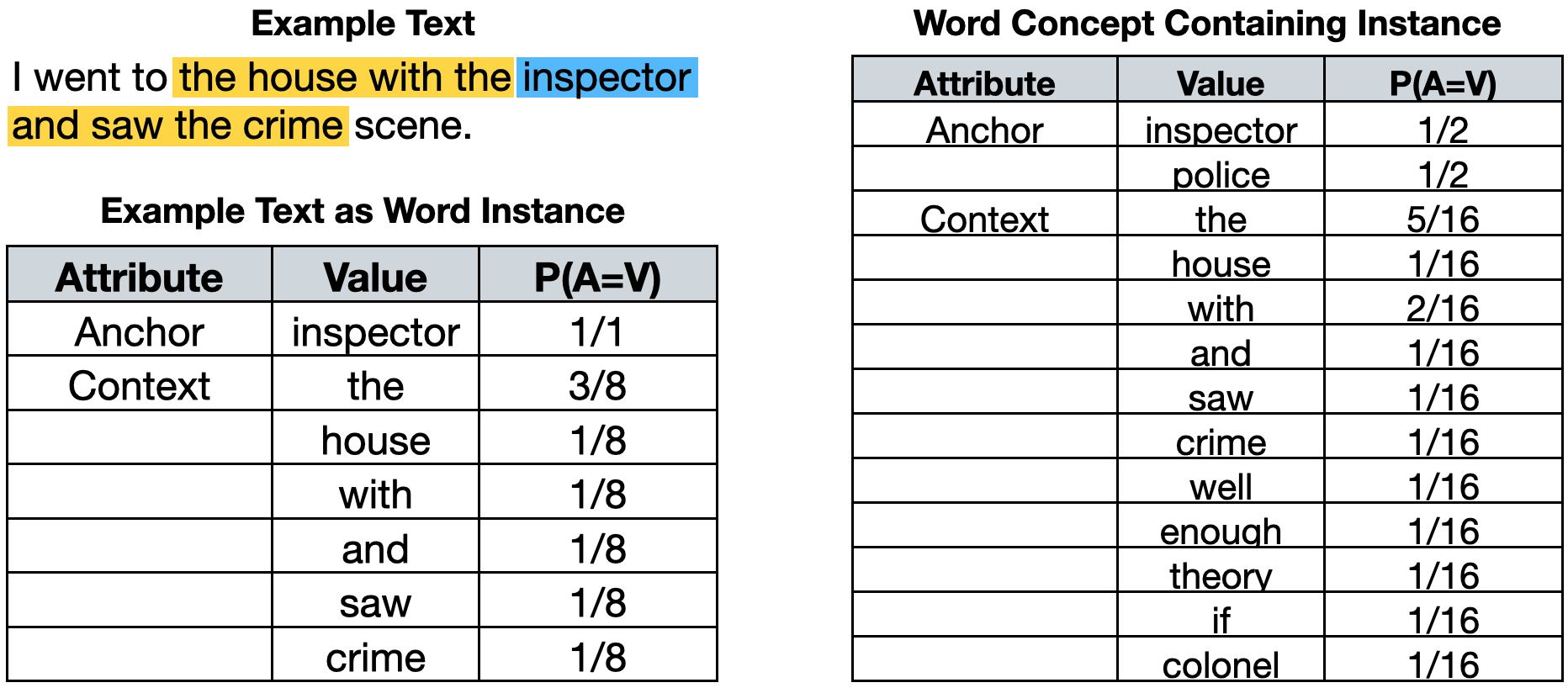}
    \vskip -0.02in
    \caption{An example sentence with the anchor word highlighted in blue and the context words highlighted in yellow. The bottom left table shows a Word instance that encodes the highlighted text from the example, whereas the table on the right is a Word concept that incorporates this instance.}
    \label{fig:word-representation}
\end{figure}

As in Word2Vec, we wanted a bag-of-words approach to encode the words in a context window around an anchor.
To support this, we modified Cobweb's instance representation to maintain count statistics over attribute values, making instances similar to concepts.
This updated representation makes it possible to specify how frequently words appear in an instance's context. 
Figure \ref{fig:word-representation} shows an example sentence, with the anchor highlighted in blue and the context highlighted in yellow, as well as what the example looks like when represented as a Word instance.
As this example shows, we track count statistics over the attribute values.
The anchor word always has a probability of 1/1 because there is always a single word in the anchor position---this makes it akin to the nominal attributes in the original Cobweb instance representation which do not maintain count statistics because their values are always implicitly 1/1. 

However, tracking for the context attribute is different; the new instance 
representation lets the Word variant track of how frequently words appear 
across the context.
In our example, ``the'' appears three times across the eight context 
words, and we can see how the instances store counts for this element.
In essence, the probability mass for each attribute value, which in the 
original Cobweb system would always be concentrated in a single value, 
can now be distributed over multiple elements.
An important feature of this altered representation is that it does not 
keep track of the order in which the context words appear.

No changes were required for the concept representation nor were any changes required to Cobweb's performance or learning mechanisms. 
However, we modified how the system updates concepts to support the new 
instance representation so that it handles concept statistics properly.
Figure \ref{fig:word-representation} shows an example of a Word concept that contains the instance for our earlier example along with one other instance.
As we can see, the denominator for the anchor values is 2, as the concept is summarizing over two instances.
However, the denominator for context values is 16 because the system has seen a total of 16 context elements over these two instances.\footnote{Sometimes instances will have fewer context words when they represent examples near the edges of the text.}

We made no changes to the way category utility is calculated, but the semantics of the context elements differ slightly from those in the original Cobweb. 
We maintain a single attribute to represent the context elements, rather than a context attribute for each context word slot.
This lets the variant ignore the order of context words.
Additionally, we can use the concept's context counts to compute a probability distribution over words that might appear in any given context slot.
We could use this probability distribution to predict the values of multiple context words because the probability distribution over each context word is the same. 
Our choice to have a single context attribute also has the side effect of weighting equally the tasks of predicting the anchor and the context, as category utility allows at most one correct guess per attribute.\footnote{We might also imagine an approach that weights the context attribute by the window size in the category utility calculations to simulate predicting each of the context words, but this seems like it would lean too heavily towards predicting the context elements over the anchor.}

\subsection{The Cobweb Leaf System}

\begin{figure}[t!]
    \centering
    \includegraphics[width=0.8\textwidth]{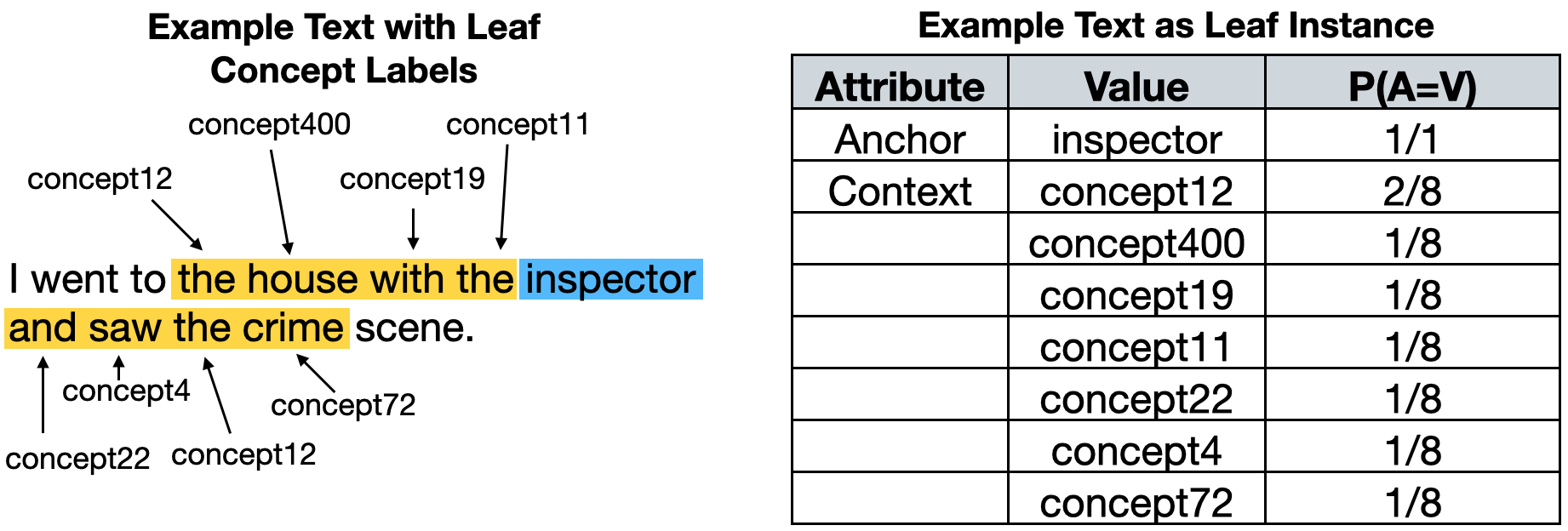}
    \caption{An annotated version of the example text showing the concept labels that are associated with each word along with the Leaf instance that describes this example. Notice, that the same word (e.g., ``the'') is not always mapped to the same concept label.}
    \label{fig:leaf-example}
    \vskip 0.20in
\end{figure}

A major insight of Word2Vec is that it represents words by their embeddings, which are in turn based on the context in which these words typically appear.
The Word system fails to take advantage of this insight, instead representing words by their unique identifiers (i.e., the word strings).
In Word2Vec, this would be akin to using a `hot-one' encoding for words rather than their embedding.
The Leaf variant aims to remedy this shortcoming by altering the Word system's representation and processing mechanisms.
Rather than representing context words by their unique strings, the Leaf approach replaces words in the instance with concept labels from its taxonomic hierarchy.
Figure \ref{fig:leaf-example} shows our example text annotated with concept labels from the Leaf system's hierarchy, with every concept label corresponding to a terminal node. 
The system does not use nonterminal nodes because they change constantly as new instances are processed---which makes them an unstable representation for later learning, so relying on leaf nodes offers greater stability. 

This new representation requires changes in the way that the Leaf system processes examples at both performance and learning time.
First, when a sequence of words is processed, it iterates over the words three times to converge on a set of concept labels.
In the first pass, it creates an instance for each word in which that word is the anchor and the context is empty, because on the first pass concept labels for adjacent words are not yet available.
At this stage, any words that have never been seen before are dropped 
because the variant has no way to represent them, even though this 
decreases the number of words in the context.
The generated instances are then categorized into the hierarchy in a non-modifying way, so that counts are not updated and no nodes are created or destroyed.
During categorization, the instance is restricted to categorization paths that contain the target anchor word.
This prevents an issue that frequently occurs for newer words where the system represents them with concepts that describe other words.
This sorting process proceeds all the way to a terminal node in the hierarchy.
The resulting concept is returned so it can be used to represent that word.
This process repeats two more times, where the instances generated for categorizing each word instead contain context labels from the previous pass, as in the example from Figure \ref{fig:leaf-example}.

This iterative process generates concept labels for each word token that take into account both the token and its surrounding context, similar to word embeddings.
However, one key difference is that Word2Vec maps every word to a single point in the embedding space.
This makes it impossible to represent homonyms---words with the same spelling but different meanings.
In contrast, our Leaf system can generate multiple nodes in the concept hierarchy to encode different senses of a word that are identified by their specific contexts.
This is highlighted in the example from Figure \ref{fig:leaf-example}, where the word ``the'' does not always map to the same concept label.\footnote{Note that ``the'' is  not a typical example of a homonym, but it serves to demonstrate the basic idea.}

Once an instance has been generated via this iterative process, performance and learning operate in the same way as for the Word variant; it tracks counts over concepts just as the other scheme tracks counts over words.
However, during preliminary testing we noted that a new leaf is created in the concept tree for every instance incorporated. 
Many of these leaves correspond to the same anchor word in slightly different contexts, with these leaves typically being grouped close to one another in the hierarchy.
Our initial system often picked an arbitrary leaf to denote a given 
context word when generating the instance description.
This makes learning relationships between anchors and their context challenging because the context features are inconsistent across instances.
To overcome this issue, we implemented a simple approach to prune groups of similar concepts into a single leaf node.
During learning, if an instance is categorized to a leaf with the same anchor word, the leaf's counts are updated instead of forking to create a new instance.
This reduces tree size drastically, with the number of leaves being 
approximately equal to the vocabulary size rather than the number of 
word tokens in the training corpus, and it improves the categorization time.

\subsection{The Cobweb Path System}

Despite the specialized concept pruning technique we developed for the Leaf system, we found that it still has a tendency to create too many concepts to represent a given word.
Further, we wanted our approach to be able to leverage the conceptual similarity of words that appear in a given context.
For example, if we have seen that a word B appears in the context of word A and we know that C is conceptually similar to B (e.g., a synonym), then we want to be able to treat C as similar to B when it appears in A's context.
This is something that is implicitly afforded by word embeddings like Word2Vec, where similar words appear close to one another in the embedding space and those behave similarly when they appear in the context of an anchor. 

Our third and final approach, the Path system, aims to support this capability.
The key insight of this variant is to represent words not just by the leaf concept they map to, but by the entire path through the hierarchy to get to that leaf.
The general idea is that if different nodes are conceptually related, than they will be grouped together in the hierarchy and have similar overlapping ancestors.
Representing context words by their conceptual paths should overcome the challenges we encountered in the leaf system and enable it to better support synonym effects in the context.

To support this capability, we extend the instance and concept representations used by the Leaf variant.
These earlier representations, shown in Figure \ref{fig:leaf-example}, contain the labels for the terminal leaves each word was categorized as.
The Path system extends the instance description by walking through the ancestors of each leaf concept and tracking count statistics on these ancestors as well.
As a result, higher concepts, which are shared by more leaves tend to have higher frequency counts.
For example, Figure \ref{fig:leaf-example} would include a value for the root concept with a count of 8/8 since it applies to every word in the context.
It is worth noting that the count statistics no longer will produce a valid probability distribution over the context elements, we will address this in a moment with an updated category utility calculation.

The performance mechanism for the Path system is nearly identical to that of the Leaf variant, but now it operates over the extended instance and concept representations that track both the terminal and nonterminal concept counts.
The biggest change to the approach was made to the learning mechanism.
The main reason that the Leaf system uses terminal concepts to represent words is that these nodes might be moved around but they are never deleted.
However, this is not the case for nonterminal concepts, which are constantly getting split/deleted and created as new instances are incorporated into the tree.
To facilitate keeping the count tables up to date, every concept now maintains a list of pointers to all of the concepts that refer to it.
If a concept is deleted, then all references to it are removed from the count tables of all the other concepts that reference it.
Similarly, if a concept is merged (which creates a new parent to that concept), then the concepts that refer to it are updated to add the new parent and its counts are updated to reflect the current concept.
This produces an effect that when two concepts that appear in a given count table are merged, a parent entry is created that ultimately has the sum of their counts.
While tracking cross links between concepts adds complexity, it presents exciting possibilities, such as enabling our system to dynamically update the conceptual representation of previously seen words.

As mentioned previously, the count statistics for context elements no longer produce a valid probability distribution. 
The key issue is that now context elements can take on multiple concept labels (for each concept along the leaf's path) rather than a single label.
To support this multi-label idea, we re-interpret the task of predicting concept labels as predicting whether or not each attribute is associated with a given context element.
This yields an updated category utility calculation between a parent $C$ and its children $C_1, ... C_n$, 

\vskip 0.05in
\begin{equation*}
\frac{\sum_{k=1}^n P(C_k) \left[ EC(C_k) - EC(C) \right]}{n} ~ , 
\end{equation*}

\noindent 
where
\vspace{-0.05in}
\begin{equation*}
    EC(C) = \sum_{A \in C_{anchors}} \left[ P(A|C)^2 \right] + \sum_{L \in C_{context}} \left[ \frac{P(L|C)^2 + P(\neg L|C)^2}{\textrm{\# concepts in tree}} \right] + 
    \frac{\textrm{\# concepts not in C}}{\textrm{\# concepts in tree}} ~ .
\end{equation*}

\vskip 0.05in
\noindent
EC provides the formula for the expected number of correct guesses that will be made given a concept C. 
The $P(A|C)^2$ applies to the anchor values and is identical to how category utility is calculated in the original Cobweb formulation. 
$P(L|C)$ and $P(\neg L|C)$ represent the probability that a concept label appears or does not appear within the context; they sum to 1.
The middle term calculates the number of concept labels that this concept has seen that we can expect to correctly guess.
The third term accounts for all of the concept labels that do not appear in the concept; we expect to correctly guess that they will not appear because they have zero probability given the concept label.
We divide the second and third terms by the number of concepts in the tree, so that the total number of expected correct guesses over all the concept labels can never exceed 1, making it have comparable weight to anchor.

% Peter said we are not doing concept pruning, so don't need to describe...
%One challenge we encountered with this approach is that the size of the count tables stored within each of the concepts can become quite large .

Combining these features lets the new system learn word representations that take into account both the word and its context.
This approach supports the ability to account for the conceptual similarity of words that appear within an anchor word's context by tracking and using path information.
It also supports the ability to dynamically update the underlying representations of words as new words are encountered without having to revisit previously seen words. 
This is in direct alignment with our goal of supporting efficient, human-like learning of language models in an incremental and continual fashion. 

\section{Experimental Evaluation}

As a basis for our evaluation, we started with all 1040 sentences from the publicly released Microsoft sentence completion challenge data set, which are extracted from Sir Arthur Conan Doyle's Sherlock Holmes stories.
We preprocessed these sentences to remove punctuation and stop words.
Finally, we removed all sentences that had a length shorter than ten words, so that we have sufficient context for categorizing words.
We leveraged the resulting 374 sentences to conduct three evaluations: synonym grouping, homonym grouping, and synonym recall.

\subsection{Synonym Grouping}
To evaluate each system's ability to group synonyms, we randomly selected 200 sentences as well as the five words that occur most frequently across these sentences (after removing stop words).
We duplicate each sentence five times and in each case we replaced any occurrences of the words in our top-five set with a variant that includes the copy number (e.g., ``door'' becomes ``door-1'', ``door-2'', \dots, ``door-5''). 
In this synthetic data set, which totals 1000 sentences, the different variations of the top-5 words effectively represent synonyms of one another because they share identical contexts.

To evaluate each approach, we built a concept hierarchy by randomly shuffling the training sentences and then sequentially incorporating them.
We analyzed the hierarchical concept organization that was produced by each approach to determine if synonyms were grouped together.
To quantitatively evaluate the groupings, we used a measure called Adjusted Rand Index (ARI) \citep{hubert1985comparing}.
This measure, which ranges from $-1$ to 1, compares clusterings to see how well they agree.
The measure corrects for chance, so a random clustering will produce a score of 0, a perfect match will produce a score of 1, and a poor match will produce a $-1$.

As a ground truth for our evaluation we labeled each instance that had a synonym variant as an anchor using the word it was based on.
The system never sees the base words, only the variants, so this is an unsupervised learning task.
The ARI measure supports comparison of two flat clusterings, it does not directly support evaluation of hierarchical clusterings. 
To get flat cluster assignments out of each approach, we successively split nodes at the root of the concept hierarchy, always splitting the node that yields the highest category utility.
After each split, we generated a flat clustering of all the instances using the cluster label just below the root.
We compared the cluster organization of the synonym variants at the root with the ground truth labels for the base synonyms using ARI.
We repeated this process, successively splitting clusters at the root until the ARI score was maximized.
This maximum score, which is a measure of the best flat clustering we might extract from our hierarchy, provides a measure of synonym grouping.
We repeated our evaluation 6 times, starting with a different set of 200 randomly selected sentences each time.

\subsection{Homonym Grouping}

Our second evaluation looked at how well our systems are able to distinguish between homonyms.
Similar to the synonym grouping test, we randomly selected 200 sentences as well as the five words that occur most frequently across these sentences (after removing stop words).
We duplicated each sentence five times, but rather than replacing the occurrences of our top-five words with variants, we replace all the other words with variants that include a copy number.
Note, if a sentence contains more than one top-five word, then each keeps their original form and are not converted into variants.
The result is a set of 1000 sentences where the top five words each appear in five distinct contexts.

To assess the groupings, we applied an approach very similar to the synonym evaluation.
However, in this case the ground truth labels for each instance with a homonym anchor is produced by appending the homonym to the number attached to all words in its context.
For example, if tor was our homonym/anchor and it appeared in the sentence, ``I-1, found-1 the-1 black-1 tor upon-1 which-1 I-1 had-1'', then the ground truth label for this instance would be ``tor-1''.
The system only ever sees the unique word (``tor''), whereas the ground truth labels identify which homonym it is (``tor-1''), so this is an unsupervised learning task.
Similar to the synonym grouping task, we converted the hierarchical organization into flat clusterings that could be evaluated with ARI and chose the clustering with the highest score.
We repeated our evaluation 6 times starting with a different set of 200 random sentences each time.

\subsection{Synonym Recall}

Our third assessment was a synonym recall test.
For this evaluation, we selected the first 400 sentences as well as the 50 top words that occur most frequently across the entire corpus (after stop word removal).
We then applied the same synonym generation process used for the synonym grouping task, which duplicates the sentences five times and replaces any occurrences of the top-50 words with variants based on the copy number.
This produces a data set containing 2000 sentences. 
Next, we randomly shuffled these sentences and categorized them using the respective system to build up a categorization tree.
Whenever we encountered a variant of one of our top-50 synonyms, we first categorize it in a non-modifying way.
Using the returned concept, we compute the probability that it would generate the anchor word or one of its synonyms.
This probability provides a measure of whether the incoming instance would be properly categorized.
We record this probability along with how many times the word or one of its synonyms has been seen before.
Finally, the instance is then incorporated into the tree in a modifying way (i.e., with learning turned on).
Using recorded probabilities and counts, we can generate learning curves that show how each system's recall of relevant concepts improves with experience.

\begin{figure}[t!]
    \centering
    \includegraphics[width=1\textwidth]{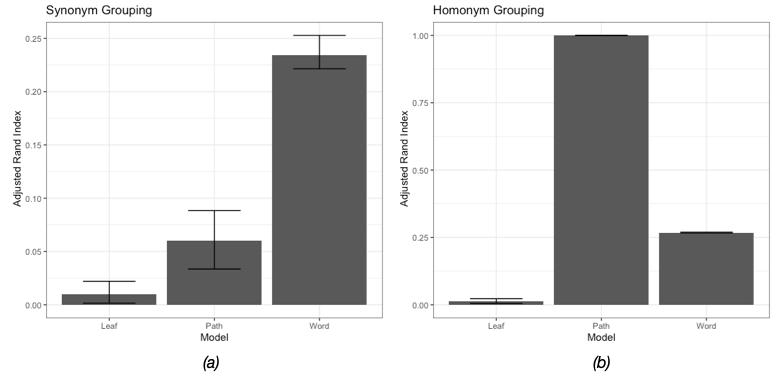}
    \vskip -0.10in
    \caption{Adjusted Rand Index score (ranges from -1 to 1, 0 is chance) on the synonym (a) and homonym (b) grouping evaluations. The bar graphs show the average across 6 runs along with 95\% confidence intervals.}
    \label{fig:synonym-grouping}
\end{figure}

We also attempted to evaluate Word2Vec on a similar task, so we can compare it to our approaches.
To provide a somewhat comparable measure, we used a Word2Vec model to predict the anchor word given the context words and summed the prediction probabilities for the anchor word and of its synonyms.
Note, this comparison somewhat advantages our approaches, which can use the anchor word as part of its instance description to retrieve the relevant concept; there is no way to make use of the anchor word in this way within Word2Vec.\footnote{We tried adding the anchor word to the  context word list, but performance was similar, so we do not report  results.}

To generate our results, we use an existing implementation of Word2Vec made available through the Gensim package \citep{gensim}.
This implementation has a number of hyperparameters that we had to choose for our evaluation.
We chose to use the CBOW approach, as this was the default for the package.
Similar to our Cobweb approaches, we set the window size to 4.
We used a embedding size of 32 and a learning rate of 0.05.
As our evaluation is an incremental learning task, for each sentence we trained Word2Vec on all previously seen sentences and applied it to predict synonyms in the current sentence (previously unseen).
For each training step, we trained the model for 100 epochs.
We then generated a prediction using the standard CBOW approach of predicting the anchor word using the words in the context window.

% Lastly in addition to these three evaluations, we also analyze the runtime of our approaches. 
% This analysis makes it possible to evaluate the viability of using our approaches for incremental and continual learning.
% Finally, it makes it possible to compare to other language modeling approaches such as Word2Vec.

%\begin{table}[t!]
%    \centering
%    \caption{Average 5-fold cross validated grouping scores and their standard deviation by model, lower is better}
%    \begin{tabular}{c|c}
%        \hline
%        Model & Avg. Grouping Score \\
%        \hline
%        Word & mean=6.32 (std=4.25) \\
%        Leaf & {\bf mean=1.77 (std=1.18)} \\
%        Path & mean=12.93 (std=3.80) \\
%        \hline
%    \end{tabular}
%    \label{tab:grouping-score}
%\end{table}

\subsection{Experimental Results}

\begin{figure}[t!]
    \centering
    \includegraphics[width=0.75\textwidth]{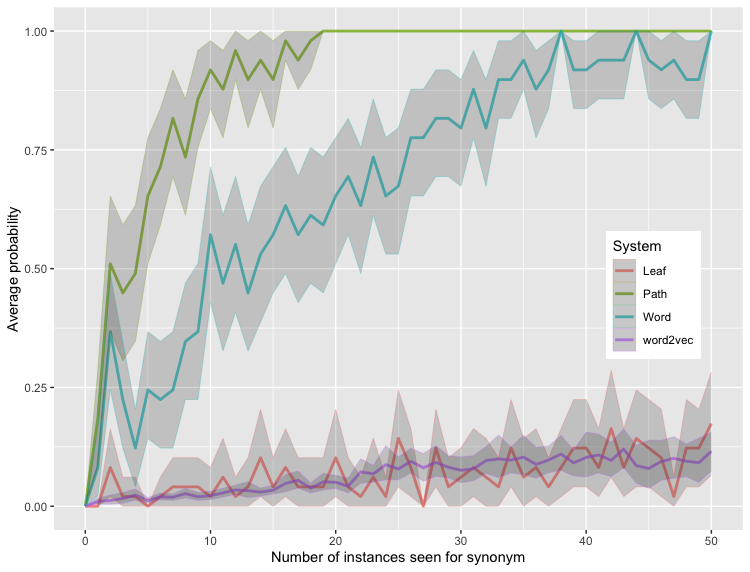}
    \caption{Average probability for the anchor word or one of its synonyms with respect to the number of previously seen instances for that word/synonym; the shaded regions denote 95\% confidence intervals.}
    \label{fig:learning-curve}
    \vskip 0.15in
\end{figure}

First, we applied our three approaches to the grouping tasks.
Figure \ref{fig:synonym-grouping} shows the average grouping scores and 95\% bootstrap confidence intervals (from 6 independent runs) for the synonym (a) and homonym (b) tasks.
These results suggest that the leaf system has close to chance performance (ARI is approximately zero) on both tasks.
However, the Path and Word variants do substantially better than chance.
On the homonym task, the Path system seems to almost perfectly separate the homonyms (ARI is approximately one). 
There appears to be a trade off, with the Path approach doing better on 
the homonym task and the Word variant doing better on the synonym task.

Next, we conducted the synonym recall test.
Figure \ref{fig:learning-curve} shows the result of this evaluation.
The Word system has good overall performance, quickly improving as the number of instances experienced increases.
The Leaf approach, on the other hand, improves very slowly.
This may be a result of the difficulty the system has with mapping the same context words to different concept labels.
The Path system has the fastest improvement; this is likely due to this approach's ability to better represent the similarities between context words using paths and to dynamically update its representation given new experiences.
Finally, we found that Word2Vec has performance that is comparable to the Leaf system.
This suggests that our approaches do at least as well as Word2Vec, with the Word and Path variants doing substantially better.

To explore the efficiency of our new approaches, we analyzed their run time during incremental training.
The original Cobweb approach has a run time of $O(N \times B^2 \times log_B(N) \times AV)$.
In the case of the Word system, the number of attribute values is directly proportional to the size of the vocabulary V.
Therefore its run time is $O(N \times B^2 \times log_B(N) \times V)$.
In contrast, the leaf approach has an attribute value for every possible anchor word (V, the vocabulary of unique words that appear in the corpus) and at most an entry for every possible leaf concept (approximated with N because every instance becomes a leaf) for the context elements.
Therefore, the leaf approach's run time is $O(N \times B^2 \times log_B(N) \times (V + N))$
Finally, the path system might have an entry for every concept in the tree, so its attribute value table is even larger.
Its run time is $O(N \times B^2 \times log_B(N) \times (V + C))$, where C is the number of nodes in the concept tree.
Table \ref{tab:runtime} summarizes these run times and compares them with the Word2Vec training times.
Looking across these run times, we can see that the Word approach is the most efficient.
It is even faster than Word2Vec, which is generally described as a very efficient language modeling system.
The Leaf system has a similar complexity to the Word2Vec (with the $N^2$ term dominating).
The Path approach also has a similar complexity.
Although C is typically larger than N, C is an upper bound for the number of concept labels stored in a node---typically this value is much smaller.

\begin{table}[t!]
    \centering
    \caption{Asymptotic time needed to incrementally train on a corpus of text.}
    \vskip 0.15in
    \begin{tabular}{c|c}
        \hline
        Approach & Asymptotic Run Time \\
        \hline
        Word2Vec-CBOW & $O(N^2 \times E \times \left( W \times D + D \times log_2(V)) \right)$ \\
        Word2Vec-Skip-Gram & $O(N^2 \times E \times \left( W \times (D + D \times log_2(V)) \right)$ \\
        Cobweb-Word & $O(N \times B^2 \times log_B(N) \times V)$ \\
        Cobweb-Leaf & $O(N \times B^2 \times log_B(N) \times (V + N))$ \\
        Cobweb-Path & $O(N \times B^2 \times log_B(N) \times (V + C))$ \\
        \hline
    \end{tabular}
    \label{tab:runtime}
\end{table}

\subsection{General Discussion}

Our results show that the Path and Word approaches are able to group synonyms and homonyms substantially better than chance, with the Path variant achieving near perfect performance on the homonym task. 
Additionally, the Path and Word systems both do quite well on the synonym recall test.
We expected that that the Path approach would yield the best performance on all three tasks and were surprised to find that the Word system also has robust performance across these tasks.
We were also surprised to see what looks like a trade off between the Path and Word approaches on the Synonym and Homonym grouping tasks.
In general, these results are promising suggesting that they are generating reasonable organizations of their experiences that support recall. 

% with the Leaf model achieving near chance performance.

A key characteristic of the Word system is that it is not learning an intermediate representation for the context words, it uses the word counts directly.
On the synonym grouping task, where many of the same combinations of words appear, we believe that the Word approach's fixed representation may yield an advantage.
However, we believe that the Path system will have better relative performance when we apply it to more natural data that is not constructed by duplicating sentences multiple times. 
This is something we should explore in future work.

The Path system achieves a near perfect score (ARI is approximately one) on the homonym grouping task, whereas the Word variant's performance is well above chance (ARI is approximately 0.25).
We suspect that the Path system's representation is responsible for this difference.
Context words for homonyms appear very infrequently.
The Word approach matches context elements by their specific strings, and the sparsity hinders learning. 
In contrast, the Path variant is able to identify similarities between different context elements (when they share paths), which makes it more robust to sparse context elements.
As a result, it is better able to distinguish homonyms.
It is worth mentioning that Word2Vec does not support the ability to represent homonyms, every word maps to a single point in the embedding.
Thus, our systems have exceeded Word2Vec in this respect.

On the synonym recall test, we find that the Path system has the most rapid progress.
We suspect this is because the system is able to adjust its representation to support learning and recall.
The downside of this dynamic representation is that it may make it harder to learn over them, especially when they are changing.
We suspect this is why we see better performance from the Path system on this task, but better performance from the Word variant on the synonym grouping tasks.
We find that the Leaf system has comparable performance to Word2Vec, and that the Word and Path approaches perform substantially better.
The amount of data we provide Word2Vec is substantially lower than what is typically used for this approach.
% The gensim Word2Vec documentation recommends millions of sentences, but we use less than 2000 sentences here. 
Our results highlight the advantage of our new Cobweb variants, suggesting some can learn more efficiently than Word2Vec.

Overall, we find that the Leaf system performs poorly.
The clusterings it produces on the synonym and homonym tasks score barely above chance.
On the synonym recall task, it continues to improve but progress is very slow.
The low performance is likely because the leaf concepts are too granular, with many of them representing the same word.
Despite this, we are optimistic that a variant of this approach that uses a more abstract concept representation for context words (i.e., non-terminals rather than terminals) will yield better results.
We are particularly interested in exploring approaches to identifying a level of representation that yields robust behavior (e.g., combines nodes that represent the same concept, but still effectively distinguishes among concepts).

Finally, our run-time analysis shows that our new approaches have the potential to be very efficient.
Most of our approaches are more efficient than Word2Vec, which is characterized as a very efficient language modeling approach \citep{mikolov2013efficient}.
The key feature of our Cobweb variants are their ability to learn incrementally from new training examples; in contrast, Word2Vec must retrain on all data (old and new) whenever new information becomes available
Although our current implementations (which are in Python) are slower than the highly optimized Word2Vec approach, they should be as (or more) efficient  than Word2Vec once they are optimized.

\vspace{-0.06in}
\section{Related Research}

% NELL

To our knowledge, there are few recent systems that learn language models while adhering to the human-like constraints proposed by  \citet{langley2022computational}.
A notable counterexample is Mitchell et~al.'s (\citeyear{mitchell2018never}) Never Ending Language Learner (NELL), which crawls the Web to acquire, in an incremental manner, an ever expanding knowledge base. 
NELL takes a top-down approach to extracting content about the world from text, whereas our approach operates from the bottom up, attempting to extract the semantics of words from their surrounding context.
We view the two approaches as complementary and one can imagine hybrid approaches that combine top-down and bottom-up processing to support continual language learning.

% SAGE
McLure et~al.'s (\citeyear{mclure2010learning}) SAGE is another incremental concept learner that shares many features with Cobweb.
We are not aware of its application to learning word meanings, but we may be able to adapt many ideas presented here, especially the Word and Leaf approaches, into that framework.
However, SAGE lacks the hierarchical organization of categories that is central to Cobweb, so it may be less straightforward to translate ideas from the Path variant, which leverages this structure. 
At the same time, McLure et~al.'s system operates over relational descriptions, which could benefit future versions of our systems. 
In general, we believe that there are many opportunities to leverage mechanisms for incremental concept formation, like those that underlie SAGE and Cobweb, as the basis for more efficient language learning.

% Pretraining
A common approach in the language modeling community, although very different from ours, starts with pre-trained language models and then tunes them for a target domain \citep{devlin2018bert}.
This amortizes the costs incurred when training large models by reusing them across domains on the hope that domain general knowledge from them will apply and transfer to new settings.
In theory, this should reduce the training data needed for new tasks.
Interestingly, a recent study by \citet{krishna2022} found that pre-training and tuning on the same data from a target domain (with no pre-trained language model) yields comparable performance to tuning a pre-trained model on the target data.
This suggests that little expertise transfers from pre-trained models and 
Krishna et~al.\ argue that using them is less beneficial than pre-training, which conditions weights for effective tuning. 
Thus, pre-trained language models may less useful than acquiring larger training sets for target domains.
Even so, performance remains proportional to number of cases from the target domain, and these data sets are often orders of magnitude larger than those used in the Cobweb studies.
Thus, we believe our distinctive approach offers exciting avenues for more efficient language learning.

%Convolutional Cobweb

Finally, we should discuss Convolutional Cobweb \citep{maclellan2022convolutional}, another variation on probabilistic concept formation that induces visual concepts.
Although this system does not acquire language models, it also represents context, although it operates over images (2D arrays) rather than text (1D sequences). 
Convolutional Cobweb classifies each pixel based on that pixel's value and the values of pixels in its surrounding context.
As in the Leaf variant, it replaces pixels with the leaf concepts to which they are sorted and uses the resulting description to classify the entire image.
One key difference is that during categorization Convolutional Cobweb replaces the leaf concept labels dynamically with ancestors that are just below the root.
This change avoids many of the issues encountered with the Leaf approach, which maintained many distinct labels to encode the same words.
There should be substantial opportunities for cross pollination between the  systems.
Thus, we plan to investigate how ideas from the current work can improve visual concept formation and how the Word, Leaf, and Path systems can incorporate ideas from Convolutional Cobweb.

% \vspace{-0.06in}
\section{Conclusion}
% \vspace{-0.01in}

In this paper, we presented three extensions to Cobweb---the Word, Leaf, and Path variants---that combine Word2Vec's ideas about word context with Cobweb's  human-like concept formation to support efficient acquisition of language models.
We investigated how well these approaches group synonyms and homonyms and we showed that, although the Leaf system does little better than chance, the Word and Path variants do reasonable jobs of placing synonyms together and keeping homonyms apart.
We also conducted a synonym recall test and found that the latter systems also do well on this task, with the Leaf approach again lagging behind. 
In addition, we evaluated Word2Vec on a task similar to synonym recall to see how it compares to our new approaches.
The preliminary results suggest that the Word and Path variants show potential for far more efficient learning than Word2Vec in that they showing better synonym recall with fewer training cases.

In summary, our research highlights promising new directions for research on incremental, human-like language learning.
The Word variant, which did better than expected, holds special potential for future efforts on language modeling. 
The Path variant also offers a novel method for dynamically updating the underlying representation during incremental learning.
We believe the current work sets the stage for innovative approaches to statistical language learning that differ considerably from current language inducers and we hope that it inspires further research on human-like learning. 
Just as Word2Vec led, over the past decade, to large language models with impressive abilities, we hope that our contextual extensions to Cobweb will develop into large-scale human-like language systems that exhibit efficient, incremental,  and continual learning.

% \newpage

\vspace{-0.01in}
\begin{acknowledgements} 

% \vspace{-0.02in}
\noindent
This research was funded in part by Award HR00112190076 from DARPA's POCUS program, Award 2112532 from NSF's AI-ALOE institute, Award W911NF2120101 from ARL's STRONG program, and ONR Grant N00014-20-1-2643. The views, opinions, and findings expressed are the authors' and should not be taken as representing official views or policies of these funding agencies.
\end{acknowledgements} 

\vspace{-0.08in}

{\parindent -10pt\leftskip 10pt\noindent
\bibliographystyle{cogsysapa}
\bibliography{contextual_cobweb}

}

% Leave a blank line before the closing brace to ensure the final 
% reference has the proper indentation. 

\end{document}